\documentclass[runningheads]{llncs}

\usepackage{eccv}

\usepackage{eccvabbrv}

\usepackage{graphicx}
\usepackage{booktabs}

\usepackage{multirow}
\usepackage{multicol}

\usepackage[accsupp]{axessibility}  

\usepackage{hyperref}

\usepackage{orcidlink}

\begin{document}

\title{Deconfounded Lifelong Learning for Autonomous Driving via Dynamic Knowledge Spaces} 

\titlerunning{DeLL}

\author{Jiayuan Du\inst{*}\orcidlink{0000-0003-1589-9111} \and
Yuebing Song\inst{*}\orcidlink{0009-0007-0463-5248} \and
Yiming Zhao\orcidlink{0000-0003-0325-4295} \and
Xianghui Pan\orcidlink{0009-0002-3831-0883} \and
Jiawei Lian\orcidlink{0000-0002-4816-7059} \and
Yuchu Lu\orcidlink{0000-0002-8875-4741} \and
Liuyi Wang\orcidlink{0000-0003-1368-0300} \and
Chengju Liu\inst{\dagger}\orcidlink{0000-0001-7543-0855} \and
Qijun Chen\inst{\dagger}\orcidlink{0000-0001-5644-1188}}

\authorrunning{J.~Du, Y.~Song et al.}

\institute{Tongji University, Shanghai, China \\
\email{\{dujiayuan, yuebing\_song, liuchengju, qjchen\}@tongji.edu.cn} \\
$^*$ Equal contributions \quad
$^\dagger$Corresponding author \\
}

\maketitle

\begin{abstract}
  End-to-End autonomous driving (E2E-AD) systems face challenges in lifelong learning, including catastrophic forgetting, difficulty in knowledge transfer across diverse scenarios, and spurious correlations between unobservable confounders and true driving intents. To address these issues, we propose DeLL, a Deconfounded Lifelong Learning framework that integrates a Dirichlet process mixture model (DPMM) with the front-door adjustment mechanism from causal inference. The DPMM is employed to construct two dynamic knowledge spaces: a trajectory knowledge space for clustering explicit driving behaviors and an implicit feature knowledge space for discovering latent driving abilities. Leveraging the non-parametric Bayesian nature of DPMM, our framework enables adaptive expansion and incremental updating of knowledge without predefining the number of clusters, thereby mitigating catastrophic forgetting. Meanwhile, the front-door adjustment mechanism utilizes the DPMM-derived knowledge as mediators to deconfound spurious correlations, such as those induced by sensor noise or environmental changes, and enhances the causal expressiveness of the learned representations. Additionally, we introduce an evolutionary trajectory decoder that enables non-autoregressive planning. To evaluate the lifelong learning performance of E2E-AD, we propose new evaluation protocols and metrics based on Bench2Drive. Extensive evaluations in the closed-loop CARLA simulator demonstrate that our framework significantly improves adaptability to new driving scenarios and overall driving performance, while effectively retaining previously acquired knowledge.
\end{abstract}

\section{Introduction}
\label{sec:intro}

\begin{figure}[!h]
\centering
\includegraphics[width=1.0\linewidth]{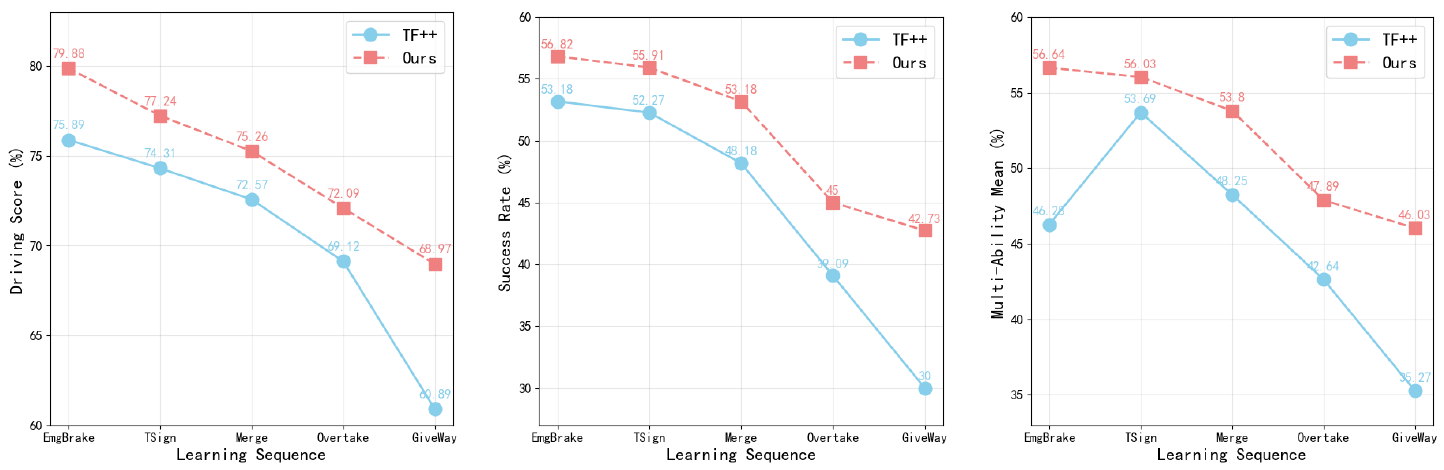}
\caption{Driving score, success rate and multi-ability success rate during the lifelong learning process. Our method not only demonstrates superior overall driving performance but also substantially mitigates catastrophic forgetting.}
\label{fig:ds_sr_mam_vs_seq}
\end{figure}

End-to-End autonomous driving (E2E-AD) methods~\cite{hu2023planning, jaeger2023hidden, Zimmerlin2024ArXiv, jia2023driveadapter, Renz2025cvpr, tang2025hip} have achieved remarkable performance in closed-loop CARLA~\cite{dosovitskiy2017carla} simulator. However, their deployment in open-world, non-stationary environments is severely hindered by catastrophic forgetting and causal confusion~\cite{chen2024end, de2019causal}. Imitation learning architectures, operating primarily as correlational engines, struggle to continuously assimilate new scenarios without overwriting historical parameters, as shown in Fig.~\ref{fig:ds_sr_mam_vs_seq}. Furthermore, treating driving as a partially observable Markov decision process reveals that these models frequently capture spurious correlations induced by unobserved confounders, leading to decision-making failures under continuous covariate shifts~\cite{ruan2024causality}.

Existing methods either improve the modeling ability through transformer-based methods and diverse auxiliary tasks~\cite{hu2023planning, jaeger2023hidden, Zimmerlin2024ArXiv, jia2025drivetransformer, tang2025hip}, or improve the interpretability of the model through interpretable intermediate representations and visualizations~\cite{chitta2021neat, shao2023safety, jia2023think}, or improve the reasoning ability and trajectory generalization performance of the model by means of large models (LLMs) and diffusion policy~\cite{yang2025drivemoe, fu2025orion, Renz2025cvpr, wang2025diffad, liao2025diffusiondrive}. However, empowering models with lifelong learning capabilities and mitigating causal confusion issues through dynamic knowledge spaces seems to be unexplored. In the closed-loop simulator CARLA~\cite{dosovitskiy2017carla}, there is also a lack of a lifelong learning benchmark.

To address aforementioned problems, we introduce \textbf{DeLL}, a \textbf{De}confounded \textbf{L}ifelong \textbf{L}earning framework for E2E-AD. The proposed architecture resolves the rigidity of fixed-capacity networks by introducing dynamic dual knowledge spaces governed by a Dirichlet process mixture model (DPMM)~\cite{LI2019128}. As a Bayesian non-parametric model, it dynamically instantiates new cluster components (knowledge anchors) as novel driving scenarios are encountered, preserving historical knowledge in isolated, specialized distributions. Our dynamic knowledge spaces alleviate catastrophic forgetting without relying on rigid task boundaries~\cite{mallya2018packnet} or computationally expensive experience replay buffers~\cite{castro2018end}. The dynamically generated knowledge anchors also serve as the mediator variables required for causal front-door adjustment~\cite{pearl2016causal, wang2024goat} in our causal feature enhancement module. We implement front-door adjustment in an attention-based way to mitigate the spurious correlations of unobservable confounders. By continuously injecting accumulated historical priors back into the network's forward propagation process, it achieves highly efficient knowledge transfer, supporting the lifelong learning goals. We also design the evolutionary trajectory decoder to match our dynamic knowledge spaces. In order to verify the effectiveness of the proposed method, we integrate the evaluation protocols of lifelong learning, design the lifelong learning task sequence based on Bench2Drive's multi-ability classification, and introduce the corresponding evaluation metrics. Our method outperforms the previous state-of-the-art in both lifelong learning and full-data learning settings.

Our contributions are summarized as follows:
\begin{itemize}
    \item A novel deconfounded lifelong learning framework ``DeLL'' for E2E-AD.

    \item DPMM-based dual knowledge spaces dynamically preserve and update latent feature and trajectory knowledge.

    \item A causal feature enhancement module leverages knowledge spaces and front-door adjustment to enhance features and alleviates spurious correlations caused by unobservable confounders.

    \item An evolutionary trajectory decoder that natively supports non-autoregressive, parallel trajectory generation.

    \item A new lifelong learning evaluation protocol based on the Bench2Drive benchmark, where our method achieves state-of-the-art closed-loop performance.
\end{itemize}

\section{Related Work}
\subsection{End-to-End Autonomous Driving in CARLA}

E2E-AD in CARLA~\cite{dosovitskiy2017carla} closed-loop simulator is predominantly built upon behavior cloning paradigms. Early works like LBC~\cite{chen2020learning} pioneers the teacher-student distillation approach, transferring privileged information to a vision-only policy. LAV~\cite{chen2022learning} expands this idea by incorporating multi-agent trajectory data to better understand social interactions. TCP~\cite{wu2022trajectory} proposes trajectory-guided control with uncertainty estimation for safer decision-making.

The advent of transformer models has profoundly reshaped the paradigm of sensor fusion. Transfuser~\cite{chitta2022transfuser} first introduces cross-modal attention mechanisms. Its successor Transfuser++~\cite{jaeger2023hidden, Zimmerlin2024ArXiv} enhances temporal modeling for occluded scenarios and long horizon planning. 
InterFuser~\cite{shao2023safety} and E2E-Parking~\cite{yang2024e2e} employ a transformer encoder for multi-modal feature fusion and a transformer decoder to generate sequential waypoints.
DriveTransformer~\cite{jia2025drivetransformer} further unifies perception and planning within transformer frameworks.

To improve the interpretability of end-to-end networks, many works introduce auxiliary tasks. Approaches such as PanT~\cite{renz2022plant}, UniAD~\cite{hu2023planning}, and VAD~\cite{jiang2023vad} emphasize joint optimization of driving sub-tasks. SimLingo~\cite{Renz2025cvpr} further integrates large language models to jointly address closed-loop driving, vision-language understanding, and language-action alignment. 
ORION~\cite{fu2025orion} integrates neural planners with symbolic guardrails to ensure strict adherence to explicit traffic rules.
ThinkTwice~\cite{jia2023think} employs a two-stage decoder with attention-based rationalization, while HiP-AD~\cite{tang2025hip} uses hierarchical predictive modeling.

Resource-efficient designs have also been explored. AD-MLP~\cite{zhai2023rethinking} replaces transformers with MLP-Mixers for real-time inference. DriveMoE~\cite{yang2025drivemoe} employs dynamic mixture-of-experts routing. DriveAdapter~\cite{jia2023driveadapter} enables plug-and-play adaptation of foundation models.
DiffAD~\cite{wang2025diffad} leverages diffusion models to address ambiguous scenarios via probabilistic sampling.

However, existing closed-loop E2E-AD methods still face fundamental challenges in incrementally learning new driving abilities in dynamic open environments as humans do. In this paper, we explore how to integrate lifelong learning in an E2E-AD framework.

\subsection{Lifelong Learning}
Lifelong learning,  also known as incremental learning, aims to enable models to acquire new knowledge while retaining previously learned information, with catastrophic forgetting being the core challenge~\cite{van2022three}. Existing approaches fall into three categories~\cite{luo2020appraisal}. Architectural strategies dynamically expand network structures: PNN~\cite{rusu2016progressive} allocates isolated columns for each task with lateral connections, while ExpertGate~\cite{aljundi2017expert} trains dedicated experts with a gating network for selection. Regularization strategies constrain parameter updates based on importance: EWC~\cite{kirkpatrick2017overcoming} penalizes changes to critical parameters via Fisher information, MAS~\cite{aljundi2018memory} computes importance from output sensitivity, LwF~\cite{li2017learning} applies knowledge distillation using old model outputs as supervision, AFA~\cite{yao2019adversarial} further introduces multi-level feature alignment losses. Rehearsal strategies revisit past data by storing exemplars or generating synthetic samples: iCaRL~\cite{castro2018end} combines distillation with prototype replay, FearNet~\cite{kemker2017fearnet} mimics biological memory with separate modules for rapid learning and long-term storage, generative models~\cite{goodfellow2020generative, kingma2013auto} produce pseudo-samples for replay~\cite{zhai2019lifelong,shin2017continual}.

Most existing lifelong learning methods focus on alleviating forgetting, while paying insufficient attention to the organization of knowledge itself. Moreover, the lifelong learning in the field of closed-loop autonomous driving in CARLA seems to be unexplored.

\section{Method}

\begin{figure}[t]
  \centering
  \includegraphics[width=1.0\linewidth]{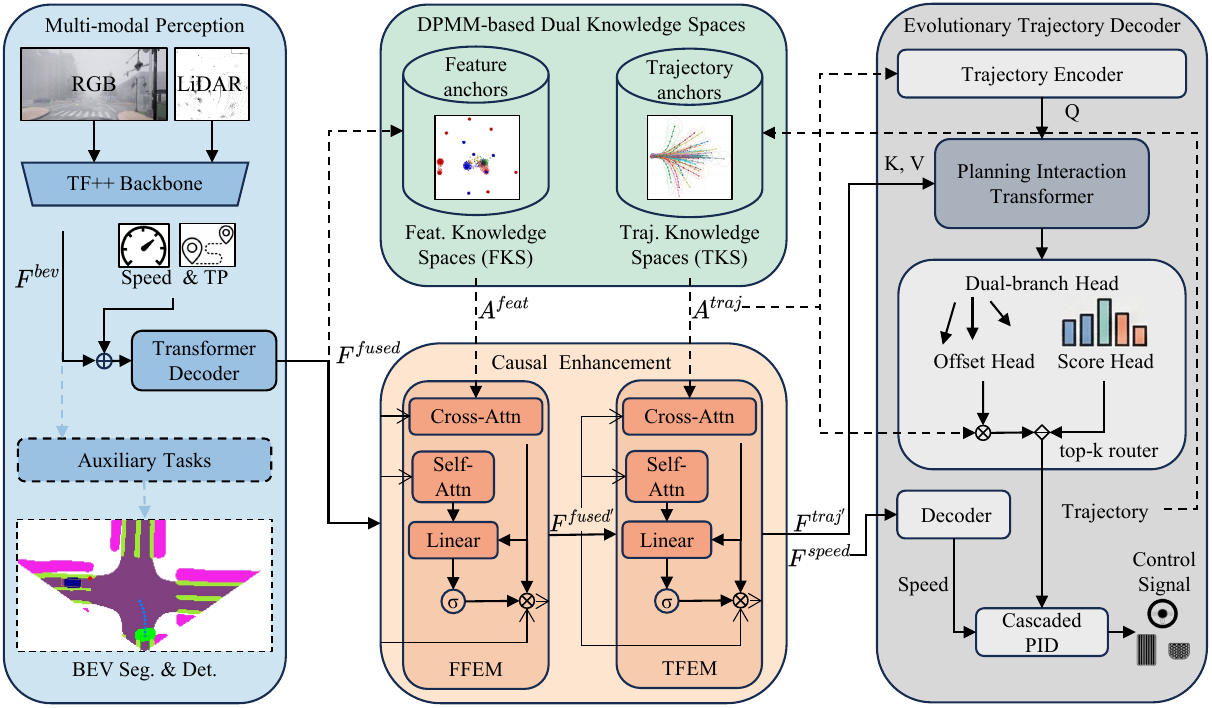}
  \caption{Overview architecture of our proposed method.}
  \label{fig:overview}
\end{figure}

We introduce DeLL, a novel deconfounded lifelong learning framework for E2E-AD. As shown in Fig.~\ref{fig:overview}, the overall workflow of the method features core modules including a multi-modal perception backbone, dynamic dual knowledge spaces, causal feature enhancement modules, and an evolutionary trajectory decoder.

\subsection{Multi-modal Perception Backbone}
The front end of the network utilizes Transfuser++~\cite{jaeger2023hidden, Zimmerlin2024ArXiv} as the base extractor,
leveraging its simplicity and modular extensibility. It first processes RGB image sequences and LiDAR point clouds independently using RegNetY~\cite{radosavovic2020designing} to obtain multi-scale features. Subsequently, cross-modal cues are fused via a multi-scale cross-attention mechanism, projecting them to generate high-dimension Bird's Eye View (BEV) feature maps $F^{bev} \in \mathbb{R}^{8 \times 8 \times 256}$.

We also introduce auxiliary learning tasks (BEV semantic segmentation and detection) to enrich the BEV representations with clear geometric and semantic boundaries. These geometrically constrained BEV features are then transformed and concatenated with the ego-vehicle's current velocity feature and target point feature. Finally, a transformer decoder equipped with 11 learnable queries pools these panoramic features into a highly compact fused scene representation vector $F^{fused} \in \mathbb{R}^{11 \times 256}$.

\subsection{Dynamic Dual Knowledge Spaces}

To overcome the memory bottleneck of fixed-capacity networks under continuous tasks, we employ the Dirichlet process mixture model (DPMM)~\cite{LI2019128} to construct explicit and implicit dual knowledge spaces in a dynamic manner. As a typical Bayesian non-parametric statistical model, DPMM allows the data to spontaneously dictate the number of cluster components, a property naturally suited to scenarios in lifelong learning where unknown data continuously flow in while the boundary is unknown. 

DPMM employs the Dirichlet process as a prior to characterize how data are generated from an infinite number of potential clusters. The Dirichlet process can be formalized as $G \sim DP(\alpha, H)$, where $G$ is a random probability measure over the parameter space, $H$ is the base distribution that represents the prior expected parameter distribution, and $\alpha$ is the concentration parameter, which controls the aggressiveness of generating new clusters. 
Expressed using the stick-breaking process, the generative process of DPMM can be represented as follows:

\begin{equation}
\theta_k \mid \lambda \sim H(\lambda), \quad \pi \mid \alpha \sim \operatorname{GEM}(\alpha), \quad v_i \mid \pi \sim \operatorname{Cat}(\pi), \quad x_i \mid v_i \sim F\left(\theta_{v_i}\right),
\end{equation} where $\theta_k$ is a latent variable drawn independently from a Dirichlet process prior, with $G$ as its base distribution. The mixing proportions $\pi$ is sampled from a generalized Ewens distribution (GEM). Variable $v_i$ assigns data point $x_i$ to a cluster and takes on the value $k$ with probability $\pi_k$, which is drawn from a categorical distribution (Cat). Each data point $x_i$ is sampled from the belonged distribution $F(\theta_{v_i})$. Clustering emerges naturally in the DPMM because observations that share the same latent parameter $\theta_k$, which is drawn from a discrete distribution, are automatically grouped together.

In our approach, we assume that the parameters of each component in the DPMM adhere to a shared Normal-Wishart base distribution. For computational efficiency, we also assume that  each active component in the DPMM can be represented as a multi-variate Gaussian with only a diagonal covariance matrix.

Our framework instantiates DPMM knowledge spaces across two different levels:

\textbf{Feature Knowledge Space (FKS)} is an implicit feature space used to cluster the fused features $F^{fused}$ extracted by the backbone network. The mission of FKS is to extract and preserve latent topological causal structures in the environment. As the learning sequence progresses, this space automatically extracts the center point of each cluster as feature knowledge anchors, denoted as $A^{feat} \in \mathbb{R}^{K_f \times 2816}$, where $K_f$ is the total number of environmental feature patterns currently identified by the DPMM.

\textbf{Trajectory Knowledge Space (TKS)} is an explicit kinematic space that directly clusters the ground truth expert trajectories from historical training data using DPMM. It constructs a physical prior action library covering maneuvers like lane changing, and sharp turns. The corresponding cluster centers are extracted as trajectory knowledge anchors, denoted as $A^{traj} \in \mathbb{R}^{K_t \times 20}$, where $K_t$ is the dynamically evolving number of trajectory prototypes.

Given the intractability of exact posterior inference in DPMM, we adopt memoVB\cite{hughes2013memoized} as an efficient approximation. memoVB decomposes global sufficient statistics into mini-batch summations, enabling online coordinate ascent updates. It leverages the nonparametric nature of DPMM to dynamically adjust cluster counts via birth and merge heuristics, helping the model escape local optima and continuously adapt to evolving data streams.
During training, the DPMM and the neural network are updated in an alternating manner.

\subsection{Causal Feature Enhancement Module}

There are unobservable confounders in the latent space that may cause spurious correlations between perception and action. To mitigate the potential influence of unobservable confounders, we introduce the causal feature enhancement module inspired by front-door adjustment~\cite{pearl2016causal, wang2024goat}. 

The core idea of front-door adjustment is to construct a front-door path between input $X$ and output $Y$ via a mediator variable $M$ ($X \rightarrow M \rightarrow Y$), and $M$ intercepts directed paths from $X$ to $Y$, while there are no unblocked back-door paths from $M$ to $Y$. Then even in the presence of unobserved confounder $U$ ($X \leftarrow U \rightarrow Y$), the interventional distribution $P(Y|do(X))$ can still be identified and calculated via the front-door adjustment formula:
\begin{equation}
    P(Y=y|do(X=x)) = \sum_m P(m|x) \sum_{x'} P(y|x', m) P(x').
\end{equation}

In our design, the sets of knowledge anchors generated by DPMM ($A^{feat}$ and $A^{traj}$) serve as the discrete state space for this observed mediator variable $M$. 
This causal intervention process is divided into two cascaded sub-modules, both adopting a unified dual-attention and gated fusion architecture.

\textbf{Fused Feature Enhancement Module (FFEM)} processes the raw multi-modal fused features $F^{fused} \in \mathbb{R}^{11 \times 256}$. First, the feature knowledge anchors $A^{feat}$ are mapped to a latent space via a projection network to obtain $\hat{F}^{fused} \in \mathbb{R}^{K_f \times 256}$. Next, a Self-attention layer extracts the internal dependencies of the current input features, followed by a cross-attention layer that uses the input features as queries (Q) and the projected knowledge anchors as keys (K) and values (V). This operation essentially calculates the expectation term $\sum_m P(m|x)$ in the formula, searching for the most fitting historical causal template for the current cluttered scene:
\begin{equation}
    F^{enhan} = \text{Attn}(Q=F^{input}, K=V=F^{knowledge}) + F^{input}.
\end{equation}

Finally, a learnable gating network built with a sigmoid-activated Multi-Layer Perceptron (MLP) adaptively calculates the fusion weight $w$, smoothly combining the raw input with the causally enhanced features:
\begin{equation}
    \begin{cases}
    F^{\text {output }}=w \odot F^{\text {enhan }}+(1-w) \odot F^{\text {input }}, \\
    w=\sigma\left[\text{MLP}\left(F^{\text {enhan }}\right)+\text{MLP}\left(\text {Attn }\left(Q=K=V=F^{\text {input }}\right)\right)\right].
    \end{cases}
\end{equation}

The FFEM ultimately outputs the deconfounded enhanced fused features $F^{fused'} \in \mathbb{R}^{11 \times 256}$.

\textbf{Trajectory Feature Enhancement Module (TFEM)} receives the subset of the FFEM output specifically responsible for trajectory prediction $F^{traj} = F^{fused'}_{1:10} \in \mathbb{R}^{10 \times 256}$. TFEM transforms the geometric coordinate anchors $A^{traj}$ from the trajectory knowledge space into high-dimensional spatiotemporal features $\hat{F}^{traj} \in \mathbb{R}^{K_t \times 10 \times 256}$ via positional encoding and temporal extension projection. Subsequently, TFEM reuses the exact same cross-attention and gated intervention mechanism, outputting trajectory features containing causal kinematic constraints $F^{traj'} \in \mathbb{R}^{10 \times 256}$. Meanwhile, the speed feature $F^{speed} = F^{fused'}_{11}$ is passed through directly to the downstream speed decoder.

\subsection{Evolutionary Trajectory Decoder}

To address the architectural rigidity of traditional fixed-channel decoders in lifelong learning, we propose an evolutionary trajectory decoder driven by the dynamic trajectory knowledge space. Leveraging the transformer’s permutation invariance and sequence flexibility, the decoder maps heterogeneous trajectory anchors $A^{traj}$ into dynamic planning token via a temporal embedding network. As the DPMM expands the cluster count $K_t$ with accumulated experience, this token pool grows naturally to achieve unbounded knowledge acquisition. These tokens then serve as queries in a planning interaction transformer, performing cross-attention against scene context features to evaluate the relevance of historical driving patterns in parallel. For trajectory generation, a dual-branch decoupled prediction head replaces traditional autoregressive methods with a parallel strategy, where a coarse-grained branch computes selection scores $Y^{logits} \in \mathbb{R}^{K_t}$ for each anchor, while a fine-grained branch predicts geometric offsets $Y^{offsets} \in \mathbb{R}^{K_t \times 20}$ to refine the predicted coordinates. The $Y^{logits}$ is then transferred into a temperature-scaled probability distribution $\hat{Y}^{probs}$. Finally, the system generates a candidate set $\hat{Y}^{trajs}$ by applying offsets to the original anchors, with the final execution output $\hat{Y}^{traj}$ selected via a Top-K routing mechanism based on $\hat{Y}^{probs}$. 
The core mathematical expression for decoding is as follows:
\begin{equation}
   \left\{\begin{array}{c}
    \hat{Y}^{trajs}=\text{MLP}\left(\operatorname{Attn}\left(Q=E\left(A^{traj}\right), K=V=F^{traj}\right)\right)+A^{traj}, \\
    \hat{Y}^{probs}=\varphi\left[\text{MLP}\left(\operatorname{Attn}\left(Q=E\left(A^{traj}\right), K=V=F^{traj}\right)\right), \tau\right], \\
    \hat{Y}^{traj}=\hat{Y}_{TopK\left(\hat{Y}^{probs}, k\right)}^{trajs},
    \end{array}\right.
\end{equation} where $\varphi[\cdot]$ is Softmax function and $\tau$ is the temperature factor.

\subsection{Loss Function}
The overall loss function is given by:
\begin{equation}
\mathcal{L}=\mathcal{L}_{sem }+\mathcal{L}_{det }+\mathcal{L}_{traj }+\mathcal{L}_{speed },
\end{equation} where the BEV semantic segmentation loss $\mathcal{L}_{sem }$ and the target speed loss $\mathcal{L}_{speed }$ are both cross-entropy losses. The BEV detection loss $\mathcal{L}_{det }$ follows the pattern of CenterNet~\cite{duan2019centernet}, which consists of heatmap loss, offset loss and size loss. The trajectory loss $\mathcal{L}_{traj }$ consists of three parts, as given below:

\begin{equation}
\mathcal{L}_{traj}=\mathcal{L}_{prob}+\mathcal{L}_{best}+\mathcal{L}_{weighted}.
\end{equation} For anchor selection probability loss $\mathcal{L}_{prob}$, we use KL divergence to minimize the difference between the predicted probability distribution and the real probability distribution, given by: 

\begin{equation}
\left\{
\begin{array}{l}
\mathcal{L}_{prob}=D_{KL}\left(Y^{probs} \| \hat{Y}^{probs}\right), \\
Y^{probs}=\varphi\left[-\left\|\hat{Y}^{trajs}-Y^{traj}\right\|_2, \tau\right],
\end{array}
\right.
\end{equation} where the ground-truth anchor selection probability distribution $Y^{probs}$ is defined by applying the softmax function to the negated distances between all predicted trajectories and the ground-truth trajectory $Y^{traj}$. The best-trajectory loss $\mathcal{L}_{best}$ and the weighted-trajectory loss $\mathcal{L}_{weighted}$ are both computed using the smooth L1 loss. The former measures the deviation between the ground truth and the closest predicted trajectory, while the latter computes the weighted sum of deviations from all anchor trajectories, with weights being the ground-truth selection probabilities. This is formulated as follows:


\begin{equation}
\left\{
\begin{array}{l}
\mathcal{L}_{best} = SmoothL1( \hat{Y}^{traj} - Y^{traj} ), \\
\mathcal{L}_{weighted} = Y^{probs} \cdot SmoothL1( \hat{Y}^{trajs} - Y^{traj} ).
\end{array}
\right.
\end{equation}

\section{Experiments}
\subsection{Dataset and Metrics}
Inspired by existing lifelong learning methods~\cite{meng2025preserving, yao2025lilodriver, lin2025h2c}, we construct a streaming data training pipeline for lifelong learning based on the Bench2Drive benchmark~\cite{Jia2024NeurIPS}. The dataset comprises approximately 512K frames covering five key driving competencies: Emergency Braking, Traffic Sign Recognition, Merging, Overtaking, and Giving Way. We leverage this inherent categorization to define five sequential learning tasks, simulating a lifelong learning scenario where tasks are encountered in order. The data volume for each competency decreases sequentially, with approximately 184K, 146K, 92K, 78K, and only 11K frames, respectively. 
To comprehensively evaluate the lifelong learning capability of models, we organize the tasks in the aforementioned order corresponding to decreasing data volume and increasing learning difficulty, thereby rigorously testing the model's resistance to forgetting and its knowledge transfer ability.

In addition to the well-established evaluation criteria in CARLA~\cite{dosovitskiy2017carla} and Bench2Drive~\cite{Jia2024NeurIPS}, we define a suite of metrics for lifelong learning that follows common practices in the field~\cite{parisi2019continual, liu2023libero, jiang2025advances}. Our framework comprises three categories: vertical metrics for temporal stability, horizontal metrics for knowledge transferability, and comprehensive metrics for overall task proficiency. While the categories are inspired by classic works~\cite{lopez2017gradient, meng2025preserving}, the exact calculations are adapted to align with the Bench2Drive benchmark’s success criteria and the nature of our tasks. Each metric is detailed in the following sections.

\textbf{Vertical Dimension (Time-Series Stability Metrics):}
\begin{itemize}
    \item \textbf{Forgetting Ratio (FR) $\downarrow$:} Measures the degree of performance decay on historically learned tasks after the model learns subsequent new tasks, relative to the performance immediately after learning them. The formula is as follows, where $SR_{i,j}$ represents the success rate on task $j$ after learning the $i$-th task. Lower values indicate stronger resistance to forgetting: \begin{equation}
F R=\frac{1}{N-1} \sum_{i=1}^{N-1}\left(S R_{i, i}-S R_{N, i}\right) / S R_{i, i}.
\end{equation}

    \item \textbf{Process Forgetting Ratio (PFR) $\downarrow$:} Captures the severity of drastic performance oscillations during the learning process caused by interference from other tasks, measuring the degree of deviation from the historical best state:
    \begin{equation}
P F R=\frac{1}{N-1} \sum_{j=1}^{N-1}\left(\frac{1}{N-j} \sum_{i=j+1}^N\left(H_{i, j}-S R_{i, j}\right) / H_{i, j}\right),
\end{equation} where $H_{i, j}=\max \left\{S R_{i, j}, i=1,2, \ldots, j-1\right\}$ represents the model's historical best performance on task $j$ after training $i$ th task.
    
\end{itemize}

\textbf{Horizontal Dimension (Knowledge Transferability Metrics):}
\begin{itemize}
    \item \textbf{Forward Transfer (FT) $\uparrow$:} Evaluates the zero-shot generalization or facilitating capability of the currently accumulated causal knowledge pool for subsequent unseen tasks:
    \begin{equation}
    F T=\frac{1}{N-1} \sum_{i=1}^{N-1}\left(\frac{1}{N-i} \sum_{j=i+1}^N S R_{i, j}\right).
    \end{equation}

    \item \textbf{Backward Transfer (BT) $\uparrow$:} Quantifies the system's average maintenance or even reverse-enhancement level across all past task sets after continuous learning:

\begin{equation}
B T=\frac{1}{N-1} \sum_{i=2}^N\left(\frac{1}{i-1} \sum_{j=1}^{i-1} S R_{i, j}\right).
\end{equation}
\end{itemize}

\textbf{Comprehensive Overall Performance Metrics:}
We also adopt the official metrics such as driving score (DS), success rate (SR) and multi-ability success rate following the CARLA~\cite{dosovitskiy2017carla} and Bench2Drive~\cite{Jia2024NeurIPS}.
\begin{itemize}
    \item \textbf{Average Driving Score $\uparrow$:} Comprehensively considers route completion and penalty deductions, reflecting the overall performance of the method.
    
    \item \textbf{Average Success Rate $\uparrow$:} The ratio of routes finished within time limits and without any traffic infractions.

    \item \textbf{Average Multi-Ability Success Rate $\uparrow$:} The average success metric derived after decoupling five different dimensions of abilities, reflecting the overall balance of the method.
\end{itemize}

\subsection{Implementation Details}
We adopt different training strategies under different settings. For full-data learning, a two-stage approach is adopted. In the first stage, we pre-train the backbone with only auxiliary task losses $\mathcal{L}_{sem }$ and $\mathcal{L}_{det }$. It is then followed by a second stage where the entire model is trained with all losses. Each stage is conducted for 30 epochs. We use a variable learning rate from 3e-4 to 3e-5. For full-data learning, our model is trained on 4 NVIDIA L40 GPUs with a total batch size of 64 for about 64 hours. The GPU memory usage of our model’s inference is approximately 1.69 GB, and the time consumption for a single forward pass is about 26.29 ms. For lifelong learning, the first task follows the aforementioned two-stage protocol, while for subsequent tasks, the backbone is frozen and training proceeds in a single stage for 30 epochs. We reset the intrinsic parameters of optimizers before initiating the training on every new incoming task.

\subsection{Main Results}

\begin{table}[h]
\caption{Lifelong learning results on Bench2Drive~\cite{Jia2024NeurIPS}. For every two rows of data, the top row is the result of baseline model and the bottom row is the result of ours.}
\label{tab:lifelong-tfpp}
\centering
\resizebox{\textwidth}{!}{
\begin{tabular}{l|cc|cccccc}
\hline
\multicolumn{1}{c|}{\multirow{2}{*}{After Task}}       & \multicolumn{2}{c|}{Overall(\%) $\uparrow$}   & \multicolumn{6}{c}{Ability(\%) $\uparrow$}                                                                      \\ \cline{2-9} 
\multicolumn{1}{c|}{}                                  & DS               & SR               & Merge  & Overtake                   & EmgBrake & GiveWay & \multicolumn{1}{c|}{TSign} & Multi-Ab Mean \\ \hline

\multirow{2}{*}{1(EmgBrake)}                           & 75.89            & 53.18            & 62.50  & 8.89                       & 40.00    & 50.00      & \multicolumn{1}{c|}{70.00} & 46.28         \\
                                                       & 79.88            & 56.82            & 43.75  & 20.00                      & 90.00    & 50.00      & \multicolumn{1}{c|}{79.47} & 56.64         \\ \cline{2-9}

\multirow{2}{*}{2(Tsign)}                              & 74.31            & 52.27            & 42.50  & 15.56                      & 83.33    & 50.00      & \multicolumn{1}{c|}{78.42} & 53.69         \\
                                                       & 77.24            & 55.91            & 52.50  & 15.56                      & 80.00    & 50.00      & \multicolumn{1}{c|}{82.11} & 56.03         \\ \cline{2-9}
                                                       
\multirow{2}{*}{3(Merge)}                              & 72.57            & 48.18            & 53.75  & 11.11                      & 51.67    & 50.00      & \multicolumn{1}{c|}{74.74} & 48.25         \\
                                                       & 75.26            & 53.18            & 55.00  & 17.76                      & 68.33    & 50.00      & \multicolumn{1}{c|}{77.89} & 53.80         \\ \cline{2-9}
                                                       
\multirow{2}{*}{4(Overtake)}                           & 69.12            & 39.09            & 37.50  & 46.67                      & 21.67    & 50.00      & \multicolumn{1}{c|}{57.37} & 42.64         \\
                                                       & 72.09            & 45.00            & 42.50  & 40.00                      & 41.67    & 50.00      & \multicolumn{1}{c|}{65.26} & 47.89         \\ \cline{2-9}
                                                       
\multirow{2}{*}{5(GiveWay)}                            & 60.89            & 30.00            & 35.00  & 8.89                       & 26.67    & 50.00      & \multicolumn{1}{c|}{55.79} & 35.27         \\
                                                       & 68.97            & 42.73            & 36.25  & 22.22                      & 51.67    & 50.00      & \multicolumn{1}{c|}{70.00} & 46.03         \\ \hline

\multicolumn{1}{c|}{\multirow{4}{*}{Lifelong Metrics}} & \multicolumn{2}{c|}{Vert. Metrics $\downarrow$} & \multicolumn{2}{c|}{Hori. Metrics $\uparrow$} & \multicolumn{4}{c}{Overall Metrics $\uparrow$}                            \\ \cline{2-9} 
\multicolumn{1}{c|}{}                                  & FR              & PFR            & FT     & \multicolumn{1}{c|}{BT}    & Avg DS   & Avg SR  & \multicolumn{2}{c}{Avg Multi-Ab SR}      \\ \cline{2-9} 
\multicolumn{1}{c|}{}                                  & 44.50           & 40.25          & 41.11  & \multicolumn{1}{c|}{52.83} & 70.55    & 44.54   & \multicolumn{2}{c}{45.23}                  \\
\multicolumn{1}{c|}{}                                  & \textbf{33.97}  & \textbf{29.8}  & \textbf{42.88}  & \multicolumn{1}{c|}{\textbf{79.63}}  & \textbf{74.69}  & \textbf{50.73}  & \multicolumn{2}{c}{\textbf{52.08}}                  \\ \hline
\end{tabular}}
\end{table}

The lifelong learning evaluation on Bench2Drive demonstrates that our framework substantially outperforms the baseline TF++~\cite{jaeger2023hidden, Zimmerlin2024ArXiv} across all sequential tasks, as shown in Table \ref{tab:lifelong-tfpp}. After the final task, our method achieves an average driving score of 74.69\%, while the average success rate improves to 50.73\%. On data-scarce `GiveWay', our model maintains 68.97\% driving score and 42.73\% success rate, versus baseline's decline to 60.89\% and 30\%. 
Our approach also reduces the process forgetting ratio from 40.25\% to 29.8\% and raises backward transfer from 52.83\% to 79.63\%, confirming that the dynamic knowledge spaces and causal intervention effectively mitigate catastrophic forgetting and enable positive knowledge transfer.

\begin{table}[h]
\caption{Comparison with lifelong learning methods adapted to E2E-AD on Bench2Drive~\cite{Jia2024NeurIPS} dataset. Bold means best.}
\label{tab:lifelong-adapt}
\centering
\resizebox{\textwidth}{!}{
\begin{tabular}{l|cc|cc|cccccc}
\hline

\multicolumn{1}{c|}{\multirow{2}{*}{Method}} & \multicolumn{2}{c|}{Training Cost $\downarrow$} & \multicolumn{2}{c|}{Vert. Metrics $\downarrow$} & \multicolumn{2}{c|}{Hori. Metrics $\uparrow$} & \multicolumn{4}{c}{Overall Metrics $\uparrow$}                            \\ \cline{2-11} 
\multicolumn{1}{c|}{}                         & Data & Time          & FR              & PFR            & FT     & \multicolumn{1}{c|}{BT}    & DS   & SR  & \multicolumn{2}{c}{Multi-Ability}      \\ \cline{1-11} 
\multicolumn{1}{c|}{Baseline~\cite{jaeger2023hidden, Zimmerlin2024ArXiv}} & $\times$ 1 & $\times$ 1     & 44.50          & 40.25          & 41.11  & \multicolumn{1}{c|}{52.83} & 70.55    & 44.54   & \multicolumn{2}{c}{45.23}  \\
\multicolumn{1}{c|}{Baseline + ER~\cite{chaudhry2019tiny}}  & $\times$ 1.35 & $\times$ 1       & 41.52         & 31.11          & 39.26  & \multicolumn{1}{c|}{79.17} & 72.80    & 47.57   & \multicolumn{2}{c}{49.86}                  \\
\multicolumn{1}{c|}{Baseline + PackNet~\cite{mallya2018packnet}}  & $\times$ 1 & $\times$ 2              & 34.11           & \textbf{25.75}          & 42.81  & \multicolumn{1}{c|}{74.50} & 74.46    & 50.65   & \multicolumn{2}{c}{51.64}                  \\
\multicolumn{1}{c|}{DeLL (Ours)}   & $\times$ \textbf{1} & $\times$ \textbf{1}  & \textbf{33.97}  & 29.80  & \textbf{42.88}  & \multicolumn{1}{c|}{\textbf{79.63}}  & \textbf{74.69}  & \textbf{50.73}  & \multicolumn{2}{c}{\textbf{52.08}}                  \\ \hline
\end{tabular}}
\end{table}

To the best of our knowledge, DeLL is the first work to perform lifelong learning in the field of closed-loop end-to-end autonomous driving (E2E-AD). Therefore, we adapt Experience Replay (ER)~\cite{chaudhry2019tiny} and PackNet~\cite{mallya2018packnet} to the 
baseline model 
for fair comparison, denoted as ``Baseline + ER'' and ``Baseline + PackNet''.
As shown in Table~\ref{tab:lifelong-adapt}, ER~\cite{chaudhry2019tiny} requires extra space to cache the dataset due to its rehearsal-based nature, while PackNet~\cite{mallya2018packnet} requires twice the training time because of the additional network pruning step. It demonstrates that our approach DeLL outperforms baselines on the vast majority of metrics, while notably requiring less training cost.

\begin{table}[htbp]
\caption{Full-data learning results of E2E-AD methods on Bench2Drive~\cite{Jia2024NeurIPS}. Bold stands for best and underlined for second best.}
\label{tab:full-data}
\centering
\resizebox{\textwidth}{!}{
\begin{tabular}{c|cc|cccccc}
\hline
\multirow{2}{*}{Method} & \multicolumn{2}{c|}{Overall(\%)$\uparrow$} & \multicolumn{6}{c}{Ability(\%)$\uparrow$} \\ \cline{2-9} 
                        & DS & SR & Merge & Overtake & EmgBrake & GiveWay & \multicolumn{1}{c|}{TSign} & Multi-Ab Mean \\ \hline
AD-MLP~\cite{zhai2023rethinking}                & 18.05 & 0.00  & 0.00  & 0.00  & 0.00  & 0.00           & \multicolumn{1}{c|}{4.35}  & 0.87  \\
TCP~\cite{wu2022trajectory}                     & 40.70 & 15.00 & 16.18 & 20.00 & 20.00 & 10.00 & \multicolumn{1}{c|}{6.69} & 14.63 \\
VAD~\cite{jiang2023vad}                     & 42.35 & 15.00 & 8.11 & 24.44 & 18.64 & 20.00 & \multicolumn{1}{c|}{19.15} & 18.07 \\
UniAD~\cite{hu2023planning}                   & 45.81 & 16.36 & 14.10 & 17.78 & 21.67 & 10.00 & \multicolumn{1}{c|}{14.21} & 15.55 \\
ThinkTwice~\cite{jia2023think}              & 62.44 & 31.23 & 27.38 & 18.42 & 35.82 & 50.00 & \multicolumn{1}{c|}{54.23} & 37.17 \\
DriveTransformer~\cite{jia2025drivetransformer} & 63.46 & 35.01 & 17.57 & 35.00 & 48.36 & 40.00  & \multicolumn{1}{c|}{52.10} & 38.60 \\
DriveAdapter~\cite{jia2023driveadapter}           & 64.22 & 33.08 & 28.82 & 26.38 & 48.76 & 50.00 & \multicolumn{1}{c|}{56.43} & 42.08 \\
DiffAD \cite{wang2025diffad}                    & 67.92 & 38.64 & 30.00 & 35.55 & 46.66 & 40.00   & \multicolumn{1}{c|}{46.32} & 38.79  \\
DriveMoE \cite{yang2025drivemoe}                & 74.22 & 48.64 & 34.67 & 40.00 & 65.45 & 40.00   & \multicolumn{1}{c|}{59.44} & 47.91  \\
ORION \cite{fu2025orion}                        & 77.74 & 54.62 & 25.00 & 71.11 & 78.33 & 30.00   & \multicolumn{1}{c|}{69.00} & 54.72  \\
TF++~\cite{jaeger2023hidden, Zimmerlin2024ArXiv}  & 84.21 & 67.27 & \underline{58.75} & 57.77 & \textbf{83.33} & 40.00 & \multicolumn{1}{c|}{\underline{82.11}} & 64.39 \\
SimLingo~\cite{Renz2025cvpr}                    & 85.07 & 67.27 & 54.01 & 57.04 & \textbf{88.33} & \underline{53.33} & \multicolumn{1}{c|}{\textbf{82.45}} & \underline{67.03} \\
HiP-AD \cite{tang2025hip}                       & \underline{86.77} & \textbf{69.09} & 50.00 & \textbf{84.44} & \textbf{83.33} & 40.00 & \multicolumn{1}{c|}{72.10} & 65.98  \\ \hline
DeLL (Ours)                                   & \textbf{86.86} & \underline{68.63} & \textbf{61.25} & \underline{62.22} & \underline{80.00} & \textbf{60.00} & \multicolumn{1}{c|}{81.05} & \textbf{68.90} \\ \hline
\end{tabular}}
\end{table}

Under the full-data training paradigm, our method achieves state-of-the-art performance among all compared end-to-end models, as shown in Table \ref{tab:full-data}. It attains the highest driving score of 86.86\% and also leads in the average multi-ability success rate with 68.9\%, reflecting well-rounded competence across diverse driving scenarios. These results confirm that the architectural innovations introduced for lifelong learning also enhance representational power and causal reasoning in static training settings.

\subsection{Ablation Study}
\begin{table}[h]
\caption{Ablation study on Bench2Drive~\cite{Jia2024NeurIPS} under lifelong learning setting}
\label{tab:lifelong-ablation}
\centering
\resizebox{\textwidth}{!}{
\begin{tabular}{cccccccc}
\hline
\multicolumn{1}{c|}{\multirow{2}{*}{Method}} & \multicolumn{3}{c|}{Overall Metrics(\%) $\uparrow$}                   & \multicolumn{2}{c|}{Vert. Metrics(\%) $\downarrow$} & \multicolumn{2}{c}{Hori. Metrics(\%) $\uparrow$} \\ \cline{2-8} 
\multicolumn{1}{c|}{}                        & Avg DS & Avg SR & \multicolumn{1}{c|}{Avg Multi-Ab SR} & FR    & \multicolumn{1}{c|}{PFR}   & BT              & FT              \\ \hline
\multicolumn{1}{c|}{Baseline Model}          & 70.55  & 44.54  & \multicolumn{1}{c|}{45.23}           & 44.50 & \multicolumn{1}{c|}{40.25} & 52.83           & 41.11           \\
\multicolumn{1}{c|}{w/o ET Dec.}             & 72.94  & 49.82  & \multicolumn{1}{c|}{50.74}           & 33.12 & \multicolumn{1}{c|}{31.76} & 72.21           & 42.98           \\
\multicolumn{1}{c|}{w/o TFEM}                & 73.00  & 48.36  & \multicolumn{1}{c|}{49.92}           & 36.43 & \multicolumn{1}{c|}{32.86} & 77.14           & 40.04           \\
\multicolumn{1}{c|}{w/o FFEM}                & 73.10  & 49.33  & \multicolumn{1}{c|}{49.02}           & 38.33 & \multicolumn{1}{c|}{30.49} & 77.32           & 36.59           \\
\multicolumn{1}{c|}{Full Model}              & 74.69  & 50.73  & \multicolumn{1}{c|}{52.08}           & 33.97 & \multicolumn{1}{c|}{29.80}  & 79.63          & 42.88           \\ \hline
\multicolumn{8}{c}{*w/o: full model without a certain module. TF++~\cite{jaeger2023hidden, Zimmerlin2024ArXiv} is used as the baseline model.}                                                                            
\end{tabular}}
\end{table}

Ablation experiments quantify the contribution of each core component within our framework, as shown in Table \ref{tab:lifelong-ablation}. 
The evolutionary trajectory decoder proves indispensable for adaptive planning. Its removal reduces the average driving score to 72.94\% and backward transfer to 72.21\%. This suggests that the dynamically expanding trajectory anchor pool continually improves driving performance and is crucial for effective knowledge preserving.
The TFEM plays a vital role in imposing causally consistent kinematic constraints. Without it, the average driving score falls to 73\%, while the process forgetting ratio rises to 32.86\%. This degradation confirms that enhancing trajectory features with front-door adjustment helps preserve maneuver-specific knowledge.
The FFEM is critical for causal feature purification, as its removal causes the forgetting ratio to increase to 38.33\% and forward transfer to drop to 36.59\%, indicating that deconfounding raw multimodal features is essential for both knowledge retention and generalization to new tasks.
The full model achieves the best performance, demonstrating that the synergistic combination of knowledge spaces and cascaded front-door adjustment is essential for robust lifelong autonomous driving.

\subsection{Visualization}

\begin{figure}[!h]
\centering
\includegraphics[width=1.0\linewidth]{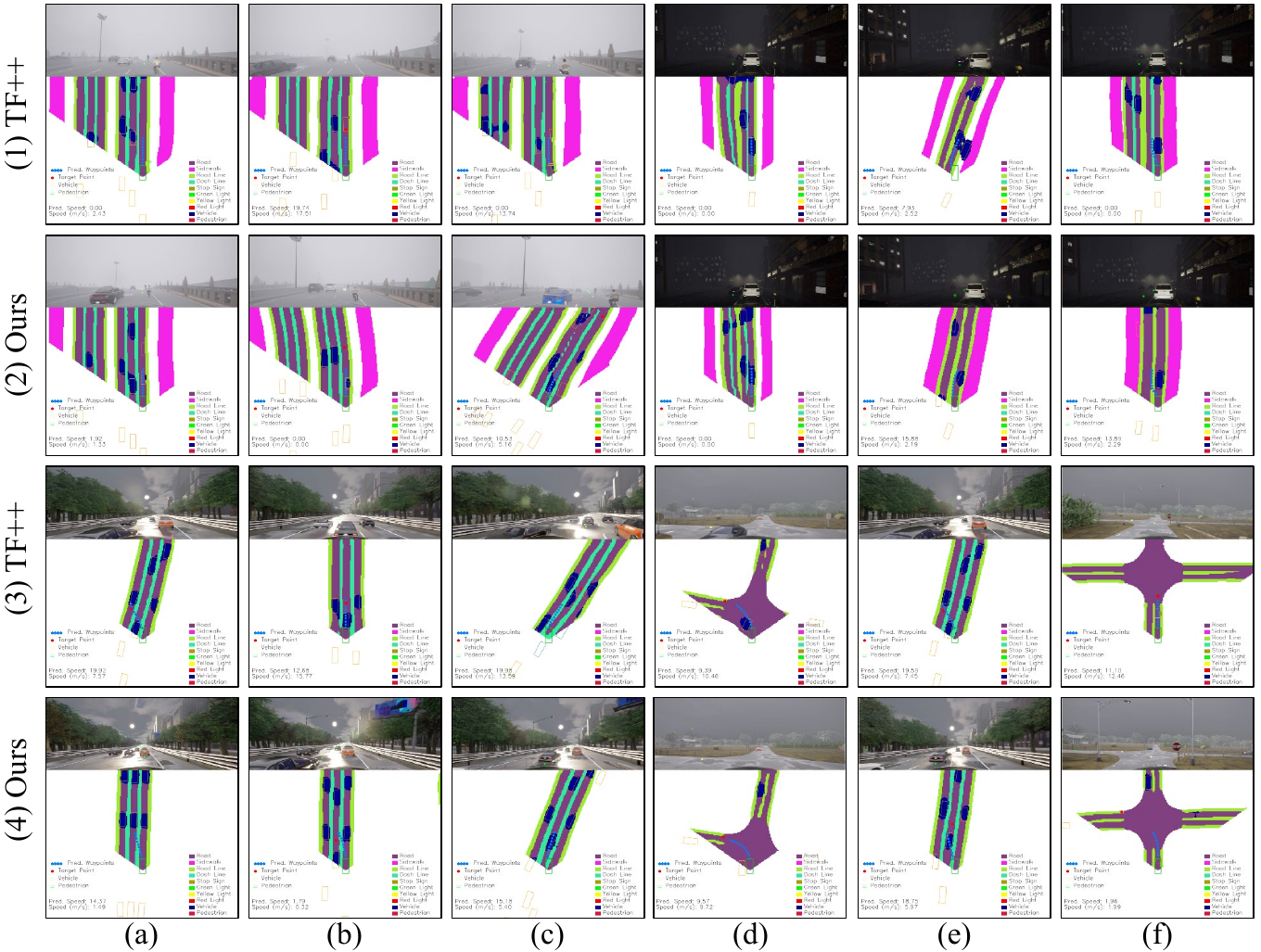}
\caption{Lifelong performance on CARLA benchmark DEV10~\cite{jia2025drivetransformer}.}
\label{fig:vis_overtake}
\end{figure}

As shown in Fig.~\ref{fig:vis_overtake}, TF++~\cite{jaeger2023hidden, Zimmerlin2024ArXiv} initially learns to decelerate and follow a slow-moving bicycle ahead after training on Task 1 (1a), but subsequent training causes it to forget this speed control strategy, leading to a collision (1b, 1c). Our method also learns timely deceleration and low-speed following initially (2a) but does not change lanes to overtake due to the absence of training data (2b). It acquires the lane-changing overtaking skill after training on Task 4 (2c).
TF++ and our method are both unable to perform parking spot exit maneuvers during the early stages of training (1d, 2d). After learning from Task 3, both methods acquire this capability (1e, 2e). However, following subsequent training, TF++ exhibits incorrect velocity predictions, potentially having learned a mistaken causal relationship between 0 velocity and a stationary vehicle ahead (1f). In contrast, our method continues to predict velocity correctly under the same conditions (2f).
TF++ cannot reliably master lane merging (3a, 3b, 3c), whereas our method is able to decisively merge by exploiting traffic gaps on (4a) and gradually learns to decelerate in dense traffic before accelerating to complete lane changes (4b, 4c), illustrating the progressive accumulation of driving knowledge.
When facing unseen scenarios at the early stage of learning, TF++'s planned trajectories often conflict with nearby vehicles (3d, 3e), while our method generates reasonable trajectories (4d, 4e), indicating forward transfer of knowledge. After learning subsequent tasks, TF++ forgets to decelerate at stop signs (3f), while our method retains this ability (4f), demonstrating resistance to forgetting.

\begin{figure}[!h]
\centering
\includegraphics[width=1.0\linewidth]{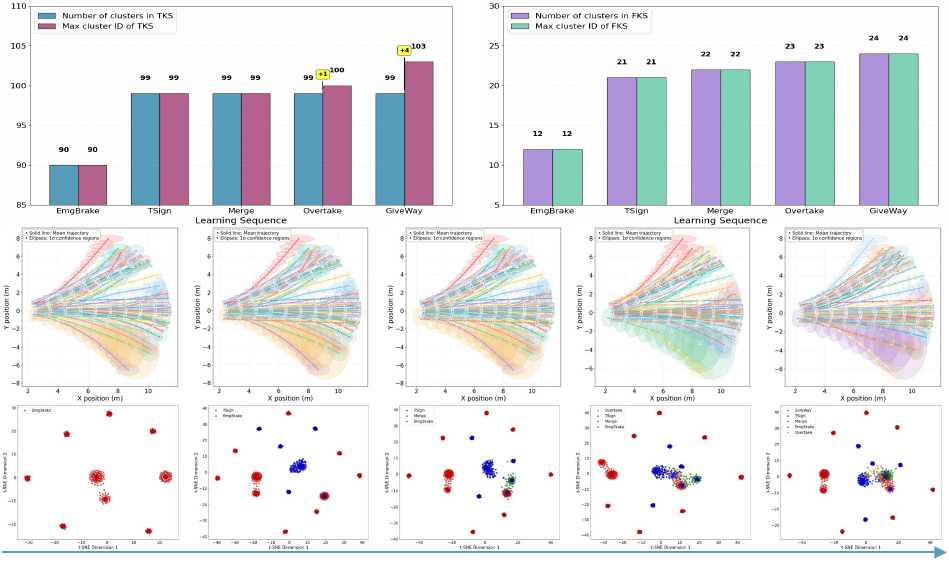}
\caption{Clustering results of dynamic knowledge spaces during the learning process.}
\label{fig:feat_knowledge_space}
\end{figure}

Fig.~\ref{fig:feat_knowledge_space} presents the clustering results of dynamic knowledge spaces during the learning process, including numbers, IDs and visualizations. In particular, we sample 50 data points per cluster from the feature knowledge space and project them into 2D using t-SNE~\cite{2008Visualizing} for visualization, with colors indicating their associated driving ability. Notably, some driving capabilities encompass multiple clusters. While most clusters are clearly separated, a small number of clusters from different capabilities exhibit spatial proximity or partial overlap, indicating coupling relationships among distinct driving abilities.

\section{Conclusion}
In this paper, we propose DeLL, a deconfounded lifelong learning framework for end-to-end autonomous driving. Our method introduces dual dynamic knowledge spaces based on DPMM to incrementally preserve and update both latent feature representations and explicit trajectory priors. By leveraging these knowledge anchors as mediators in a causal front-door adjustment mechanism, the framework effectively mitigates spurious correlations caused by unobserved confounders. Additionally, an evolutionary trajectory decoder enables non-autoregressive, parallel planning. Extensive experiments in CARLA demonstrate that DeLL achieves superior driving performance, significantly reduces catastrophic forgetting, and enhances knowledge transfer across sequential tasks.

Despite its effectiveness, our method has certain limitations. The alternating optimization between DPMM updates and network training introduces computational overhead and training cost. In addition, the current framework operates within simulated environments, so the domain gap between CARLA and real-world deployment remains an open challenge.


\section*{Acknowledgements}
This paper is supported by the National Natural Science Foundation of China under Grants No. 62473295, No. 62233013 and No. 62333017. 

We thank Zhiyong Bao (2353604@tongji.edu.cn) for his contribution to the Comparison with Adaptation of Mainstream Lifelong Learning Methods and for his help in building the lifelong learning benchmark on the CARLA simulator for end-to-end autonomous driving.
%
%
\bibliographystyle{splncs04}
\bibliography{main}

@String(CVPR  = {IEEE Conf. Comput. Vis. Pattern Recog.})

@String(ICCV  = {Int. Conf. Comput. Vis.})

@String(ECCV  = {Eur. Conf. Comput. Vis.})

@String(NeurIPS = {Adv. Neural Inform. Process. Syst.})

@String(ICLR  = {Int. Conf. Learn. Represent.})

@String(CVPR  = {CVPR})

@String(ICCV  = {ICCV})

@String(ECCV  = {ECCV})

@String(NeurIPS = {NeurIPS})

@String(ICLR  = {ICLR})

@inproceedings{dosovitskiy2017carla,
  title={{CARLA}: An open urban driving simulator},
  author={Dosovitskiy, Alexey and Ros, German and Codevilla, Felipe and Lopez, Antonio and Koltun, Vladlen},
  booktitle={CoRL},
  pages={1--16},
  year={2017},
  organization={PMLR}
}

@inproceedings{Jia2024NeurIPS,
  title={Bench2{D}rive: Towards Multi-Ability Benchmarking of Closed-Loop End-To-End Autonomous Driving},
  author={Xiaosong Jia and Zhenjie Yang and Qifeng Li and Zhiyuan Zhang and Junchi Yan},
  booktitle={NeurIPS},
  year={2024}
}

@book{pearl2016causal,
  title={Causal inference in statistics: A primer},
  author={Pearl, Judea and Glymour, Madelyn and Jewell, Nicholas P},
  year={2016},
  publisher={John Wiley \& Sons}
}

@article{chen2024end,
  title={End-to-end autonomous driving: Challenges and frontiers},
  author={Chen, Li and Wu, Penghao and Chitta, Kashyap and Jaeger, Bernhard and Geiger, Andreas and Li, Hongyang},
  journal={TPAMI},
  volume={46},
  number={12},
  pages={10164--10183},
  year={2024},
  publisher={IEEE}
}

@inproceedings{de2019causal,
  title={Causal confusion in imitation learning},
  author={De Haan, Pim and Jayaraman, Dinesh and Levine, Sergey},
  booktitle={NeurIPS},
  volume={32},
  year={2019}
}

@book{ruan2024causality,
  title={When Causality Meets Autonomy: Causal Imitation Learning to Unravel Unobserved Influences in Autonomous Driving Decision-Making},
  author={Ruan, Kangrui},
  year={2024},
  publisher={Columbia University}
}

@inproceedings{hughes2013memoized,
  title={Memoized online variational inference for Dirichlet process mixture models},
  author={Hughes, Michael C and Sudderth, Erik},
  booktitle={NeurIPS},
  volume={26},
  year={2013}
}

@article{meng2025preserving,
  title={Preserving and combining knowledge in robotic lifelong reinforcement learning},
  author={Meng, Yuan and Bing, Zhenshan and Yao, Xiangtong and Chen, Kejia and Huang, Kai and Gao, Yang and Sun, Fuchun and Knoll, Alois},
  journal={Nature Machine Intelligence},
  pages={1--14},
  year={2025},
  publisher={Nature Publishing Group UK London}
}

@article{van2022three,
  title={Three types of incremental learning},
  author={Van de Ven, Gido M and Tuytelaars, Tinne and Tolias, Andreas S},
  journal={Nature Machine Intelligence},
  volume={4},
  number={12},
  pages={1185--1197},
  year={2022},
  publisher={Nature Publishing Group UK London}
}

@article{jiang2025advances,
  title={Advances in continual learning: A comprehensive review},
  author={Jiang, Mengjuan and Fan, Jiaqing and Li, Fanzhang},
  journal={Expert Systems with Applications},
  volume={294},
  pages={128739},
  year={2025},
  publisher={Elsevier}
}

@article{parisi2019continual,
  title={Continual lifelong learning with neural networks: A review},
  author={Parisi, German I and Kemker, Ronald and Part, Jose L and Kanan, Christopher and Wermter, Stefan},
  journal={Neural Networks},
  volume={113},
  pages={54--71},
  year={2019},
  publisher={Elsevier}
}

@inproceedings{liu2023libero,
  title={Libero: Benchmarking knowledge transfer for lifelong robot learning},
  author={Liu, Bo and Zhu, Yifeng and Gao, Chongkai and Feng, Yihao and Liu, Qiang and Zhu, Yuke and Stone, Peter},
  booktitle={NeurIPS},
  volume={36},
  pages={44776--44791},
  year={2023}
}

@inproceedings{lopez2017gradient,
  title={Gradient episodic memory for continual learning},
  author={Lopez-Paz, David and Ranzato, Marc'Aurelio},
  booktitle={NeurIPS},
  volume={30},
  year={2017}
}

@article{rusu2016progressive,
  title={Progressive neural networks},
  author={Rusu, Andrei A and Rabinowitz, Neil C and Desjardins, Guillaume and Soyer, Hubert and Kirkpatrick, James and Kavukcuoglu, Koray and Pascanu, Razvan and Hadsell, Raia},
  journal={arXiv preprint arXiv:1606.04671},
  year={2016}
}

@inproceedings{aljundi2017expert,
  title={Expert gate: Lifelong learning with a network of experts},
  author={Aljundi, Rahaf and Chakravarty, Punarjay and Tuytelaars, Tinne},
  booktitle={CVPR},
  pages={3366--3375},
  year={2017}
}

@article{kirkpatrick2017overcoming,
  title={Overcoming catastrophic forgetting in neural networks},
  author={Kirkpatrick, James and Pascanu, Razvan and Rabinowitz, Neil and Veness, Joel and Desjardins, Guillaume and Rusu, Andrei A and Milan, Kieran and Quan, John and Ramalho, Tiago and Grabska-Barwinska, Agnieszka and others},
  journal={Proceedings of the national academy of sciences},
  volume={114},
  number={13},
  pages={3521--3526},
  year={2017},
  publisher={National Academy of Sciences}
}

@inproceedings{aljundi2018memory,
  title={Memory aware synapses: Learning what (not) to forget},
  author={Aljundi, Rahaf and Babiloni, Francesca and Elhoseiny, Mohamed and Rohrbach, Marcus and Tuytelaars, Tinne},
  booktitle={ECCV},
  pages={139--154},
  year={2018}
}

@article{li2017learning,
  title={Learning without forgetting},
  author={Li, Zhizhong and Hoiem, Derek},
  journal={TPAMI},
  volume={40},
  number={12},
  pages={2935--2947},
  year={2017},
  publisher={IEEE}
}

@article{yao2019adversarial,
  title={Adversarial feature alignment: Avoid catastrophic forgetting in incremental task lifelong learning},
  author={Yao, Xin and Huang, Tianchi and Wu, Chenglei and Zhang, Rui-Xiao and Sun, Lifeng},
  journal={Neural computation},
  volume={31},
  number={11},
  pages={2266--2291},
  year={2019},
  publisher={MIT Press}
}

@inproceedings{castro2018end,
  title={End-to-end incremental learning},
  author={Castro, Francisco M and Mar{\'\i}n-Jim{\'e}nez, Manuel J and Guil, Nicol{\'a}s and Schmid, Cordelia and Alahari, Karteek},
  booktitle={ECCV},
  pages={233--248},
  year={2018}
}

@inproceedings{kemker2017fearnet,
  title={FearNet: Brain-Inspired Model for Incremental Learning},
  author={Kemker, Ronald and Kanan, Christopher},
  booktitle={ICLR},
  year={2018}
}

@article{goodfellow2020generative,
  title={Generative adversarial networks},
  author={Goodfellow, Ian and Pouget-Abadie, Jean and Mirza, Mehdi and Xu, Bing and Warde-Farley, David and Ozair, Sherjil and Courville, Aaron and Bengio, Yoshua},
  journal={Communications of the ACM},
  volume={63},
  number={11},
  pages={139--144},
  year={2020},
  publisher={ACM New York, NY, USA}
}

@article{kingma2013auto,
  title={Auto-encoding variational bayes},
  author={Kingma, Diederik P and Welling, Max},
  journal={arXiv preprint arXiv:1312.6114},
  year={2013}
}

@inproceedings{zhai2019lifelong,
  title={Lifelong gan: Continual learning for conditional image generation},
  author={Zhai, Mengyao and Chen, Lei and Tung, Frederick and He, Jiawei and Nawhal, Megha and Mori, Greg},
  booktitle={ICCV},
  pages={2759--2768},
  year={2019}
}

@inproceedings{shin2017continual,
  title={Continual learning with deep generative replay},
  author={Shin, Hanul and Lee, Jung Kwon and Kim, Jaehong and Kim, Jiwon},
  booktitle={NeurIPS},
  volume={30},
  year={2017}
}

@article{luo2020appraisal,
  title={An appraisal of incremental learning methods},
  author={Luo, Yong and Yin, Liancheng and Bai, Wenchao and Mao, Keming},
  journal={Entropy},
  volume={22},
  number={11},
  pages={1190},
  year={2020},
  publisher={MDPI}
}

@inproceedings{duan2019centernet,
  title={Centernet: Keypoint triplets for object detection},
  author={Duan, Kaiwen and Bai, Song and Xie, Lingxi and Qi, Honggang and Huang, Qingming and Tian, Qi},
  booktitle={ICCV},
  pages={6569--6578},
  year={2019}
}

@inproceedings{chen2020learning,
  title={Learning by cheating},
  author={Chen, Dian and Zhou, Brady and Koltun, Vladlen and Kr{\"a}henb{\"u}hl, Philipp},
  booktitle={CoRL},
  pages={66--75},
  year={2020},
  organization={PMLR}
}

@inproceedings{chen2022learning,
  title={Learning from all vehicles},
  author={Chen, Dian and Kr{\"a}henb{\"u}hl, Philipp},
  booktitle={CVPR},
  pages={17222--17231},
  year={2022}
}

@inproceedings{wu2022trajectory,
  title={Trajectory-guided control prediction for end-to-end autonomous driving: A simple yet strong baseline},
  author={Wu, Penghao and Jia, Xiaosong and Chen, Li and Yan, Junchi and Li, Hongyang and Qiao, Yu},
  booktitle={NeurIPS},
  volume={35},
  pages={6119--6132},
  year={2022}
}

@article{chitta2022transfuser,
  title={Transfuser: Imitation with transformer-based sensor fusion for autonomous driving},
  author={Chitta, Kashyap and Prakash, Aditya and Jaeger, Bernhard and Yu, Zehao and Renz, Katrin and Geiger, Andreas},
  journal={TPAMI},
  volume={45},
  number={11},
  pages={12878--12895},
  year={2022},
  publisher={IEEE}
}

@inproceedings{jaeger2023hidden,
  title={Hidden biases of end-to-end driving models},
  author={Jaeger, Bernhard and Chitta, Kashyap and Geiger, Andreas},
  booktitle={ICCV},
  pages={8240--8249},
  year={2023}
}

@article{Zimmerlin2024ArXiv,
  title={Hidden Biases of End-to-End Driving Datasets},
  author={Julian Zimmerlin and Jens Beißwenger and Bernhard Jaeger and Andreas Geiger and Kashyap Chitta},
  journal={arXiv preprint arXiv:2412.09602},
  year={2024}
}

@InProceedings{Renz2025cvpr,
  title={SimLingo: Vision-Only Closed-Loop Autonomous Driving with Language-Action Alignment},
  author={Renz, Katrin and Chen, Long and Arani, Elahe and Sinavski, Oleg},
  booktitle={CVPR},
  year={2025}
}

@inproceedings{shao2023safety,
  title={Safety-enhanced autonomous driving using interpretable sensor fusion transformer},
  author={Shao, Hao and Wang, Letian and Chen, Ruobing and Li, Hongsheng and Liu, Yu},
  booktitle={CoRL},
  pages={726--737},
  year={2023},
  organization={PMLR}
}

@inproceedings{yang2024e2e,
  title={E2e parking: Autonomous parking by the end-to-end neural network on the carla simulator},
  author={Yang, Yunfan and Chen, Denglong and Qin, Tong and Mu, Xiangru and Xu, Chunjing and Yang, Ming},
  booktitle={2024 IEEE Intelligent Vehicles Symposium (IV)},
  pages={2375--2382},
  year={2024},
  organization={IEEE}
}

@inproceedings{jia2025drivetransformer,
 author = {Jia, Xiaosong and You, Junqi and Zhang, Zhiyuan and Yan, Junchi},
 booktitle = {ICLR},
 pages = {67227--67243},
 title = {DriveTransformer: Unified Transformer for Scalable End-to-End Autonomous Driving},
 year = {2025}
}

@inproceedings{hu2023planning,
  title={Planning-oriented autonomous driving},
  author={Hu, Yihan and Yang, Jiazhi and Chen, Li and Li, Keyu and Sima, Chonghao and Zhu, Xizhou and Chai, Siqi and Du, Senyao and Lin, Tianwei and Wang, Wenhai and others},
  booktitle={CVPR},
  pages={17853--17862},
  year={2023}
}

@inproceedings{renz2022plant,
    author       = {Katrin Renz and Kashyap Chitta and Otniel-Bogdan Mercea and A. Sophia Koepke and Zeynep Akata and Andreas Geiger},
    title        = {PlanT: Explainable Planning Transformers via Object-Level Representations},
    booktitle    = {CoRL},
    year         = {2022}
}

@inproceedings{jiang2023vad,
  title={Vad: Vectorized scene representation for efficient autonomous driving},
  author={Jiang, Bo and Chen, Shaoyu and Xu, Qing and Liao, Bencheng and Chen, Jiajie and Zhou, Helong and Zhang, Qian and Liu, Wenyu and Huang, Chang and Wang, Xinggang},
  booktitle={ICCV},
  pages={8340--8350},
  year={2023}
}

@inproceedings{jia2023think,
  title={Think twice before driving: Towards scalable decoders for end-to-end autonomous driving},
  author={Jia, Xiaosong and Wu, Penghao and Chen, Li and Xie, Jiangwei and He, Conghui and Yan, Junchi and Li, Hongyang},
  booktitle={CVPR},
  pages={21983--21994},
  year={2023}
}

@inproceedings{fu2025orion,
  title={Orion: A holistic end-to-end autonomous driving framework by vision-language instructed action generation},
  author={Fu, Haoyu and Zhang, Diankun and Zhao, Zongchuang and Cui, Jianfeng and Liang, Dingkang and Zhang, Chong and Zhang, Dingyuan and Xie, Hongwei and Wang, Bing and Bai, Xiang},
  booktitle={ICCV},
  pages={24823--24834},
  year={2025}
}

@inproceedings{tang2025hip,
  title={Hip-ad: Hierarchical and multi-granularity planning with deformable attention for autonomous driving in a single decoder},
  author={Tang, Yingqi and Xu, Zhuoran and Meng, Zhaotie and Cheng, Erkang},
  booktitle={ICCV},
  pages={25605--25615},
  year={2025}
}

@article{zhai2023rethinking,
  title={Rethinking the open-loop evaluation of end-to-end autonomous driving in nuscenes},
  author={Zhai, Jiang-Tian and Feng, Ze and Du, Jinhao and Mao, Yongqiang and Liu, Jiang-Jiang and Tan, Zichang and Zhang, Yifu and Ye, Xiaoqing and Wang, Jingdong},
  journal={arXiv preprint arXiv:2305.10430},
  year={2023}
}

@inproceedings{yang2025drivemoe,
  title={Drivemoe: Mixture-of-experts for vision-language-action model in end-to-end autonomous driving},
  author={Yang, Zhenjie and Chai, Yilin and Jia, Xiaosong and Li, Qifeng and Shao, Yuqian and Zhu, Xuekai and Su, Haisheng and Yan, Junchi},
  booktitle={CVPR},
  pages={10678--10688},
  year={2026}
}

@inproceedings{jia2023driveadapter,
  title={Driveadapter: Breaking the coupling barrier of perception and planning in end-to-end autonomous driving},
  author={Jia, Xiaosong and Gao, Yulu and Chen, Li and Yan, Junchi and Liu, Patrick Langechuan and Li, Hongyang},
  booktitle={ICCV},
  pages={7953--7963},
  year={2023}
}

@article{wang2025diffad,
  title={DiffAD: A Unified Diffusion Modeling Approach for Autonomous Driving},
  author={Wang, Tao and Zhang, Cong and Qu, Xingguang and Li, Kun and Liu, Weiwei and Huang, Chang},
  journal={arXiv preprint arXiv:2503.12170},
  year={2025}
}

@inproceedings{liao2025diffusiondrive,
  title={Diffusiondrive: Truncated diffusion model for end-to-end autonomous driving},
  author={Liao, Bencheng and Chen, Shaoyu and Yin, Haoran and Jiang, Bo and Wang, Cheng and Yan, Sixu and Zhang, Xinbang and Li, Xiangyu and Zhang, Ying and Zhang, Qian and others},
  booktitle={CVPR},
  pages={12037--12047},
  year={2025}
}

@inproceedings{radosavovic2020designing,
  title={Designing network design spaces},
  author={Radosavovic, Ilija and Kosaraju, Raj Prateek and Girshick, Ross and He, Kaiming and Doll{\'a}r, Piotr},
  booktitle={CVPR},
  pages={10428--10436},
  year={2020}
}

@inproceedings{wang2024goat,
  title={Vision-and-language navigation via causal learning},
  author={Wang, Liuyi and He, Zongtao and Dang, Ronghao and Shen, Mengjiao and Liu, Chengju and Chen, Qijun},
  booktitle={CVPR},
  pages={13139--13150},
  year={2024}
}

@article{2008Visualizing,
  title={Visualizing Data using t-SNE},
  author={Laurens, Van Der Maaten and Hinton, Geoffrey},
  journal={Journal of Machine Learning Research},
  volume={9},
  number={2605},
  pages={2579-2605},
  year={2008}
}

@article{yao2025lilodriver,
  title={Lilodriver: A lifelong learning framework for closed-loop motion planning in long-tail autonomous driving scenarios},
  author={Yao, Huaiyuan and Li, Pengfei and Jin, Bu and Zheng, Yupeng and Liu, An and Mu, Lisen and Su, Qing and Zhang, Qian and Chen, Yilun and Li, Peng},
  journal={arXiv preprint arXiv:2505.17209},
  year={2025}
}

@article{lin2025h2c,
  author={Lin, Yunlong and Li, Zirui and Du, Guodong and Zhao, Xiaocong and Gong, Cheng and Wang, Xinwei and Lu, Chao and Gong, Jianwei},
  journal={TITS}, 
  title={H2C: Hippocampal Circuit-Inspired Continual Learning for Lifelong Trajectory Prediction in Autonomous Driving}, 
  year={2026},
  volume={},
  number={},
  pages={1-18},}

@inproceedings{chitta2021neat,
  title={Neat: Neural attention fields for end-to-end autonomous driving},
  author={Chitta, Kashyap and Prakash, Aditya and Geiger, Andreas},
  booktitle={ICCV},
  pages={15793--15803},
  month= {Oct.},
  year={2021}
}

@article{LI2019128,
title = {A tutorial on Dirichlet process mixture modeling},
journal = {Journal of Mathematical Psychology},
volume = {91},
pages = {128-144},
year = {2019},
issn = {0022-2496},
author = {Yuelin Li and Elizabeth Schofield and Mithat Gönen},
}

@article{chaudhry2019tiny,
  title={On tiny episodic memories in continual learning},
  author={Chaudhry, Arslan and Rohrbach, Marcus and Elhoseiny, Mohamed and Ajanthan, Thalaiyasingam and Dokania, Puneet K and Torr, Philip HS and Ranzato, Marc'Aurelio},
  journal={arXiv preprint arXiv:1902.10486},
  year={2019}
}

@inproceedings{mallya2018packnet,
  title={Packnet: Adding multiple tasks to a single network by iterative pruning},
  author={Mallya, Arun and Lazebnik, Svetlana},
  booktitle={CVPR},
  pages={7765--7773},
  year={2018}
}

\end{document}